\documentclass[lettersize,journal]{IEEEtran}
\usepackage{amsmath,amssymb,amsfonts}
\usepackage{algorithm}
\usepackage{algpseudocode}
\usepackage{array}
\usepackage{multirow}
\usepackage{makecell}
\usepackage{booktabs}
\usepackage{diagbox}
\usepackage{colortbl}
\usepackage{threeparttable}
\usepackage{pbox}
\usepackage{graphicx}
\usepackage[caption=false,font=normalsize,labelfont=sf,textfont=sf]{subfig}
\usepackage{epstopdf}
\usepackage{pdfpages}
\usepackage{picinpar}
\usepackage{textcomp}
\usepackage{xcolor}
\usepackage[table]{xcolor}
\usepackage{color}
\usepackage{soul}
\usepackage[normalem]{ulem}
\useunder{\uline}{\ul}{}
\usepackage{pifont}
\usepackage{bbding}
\usepackage{alltt}
\usepackage{verbatim}
\usepackage{siunitx}
\usepackage{stfloats}
\usepackage{flushend}
\usepackage[switch]{lineno}
\usepackage{url}
\usepackage{breakurl}
\usepackage[hidelinks]{hyperref}
\usepackage{cite}
\usepackage{xcolor}

\begin{document}
\title{PanoAir: A Panoramic Visual-Inertial SLAM with Cross-Time Real-World UAV Dataset}
\author{Yiyang Wu, Xiaohu Zhang, Yanjin Du, Tongsu Zhang, Chujun Li, Siyang Chen, Guoyi Zhang, \\Xiangpeng Xu
\thanks{Manuscript received Month xx, 2xxx; revised Month xx, xxxx; accepted Month x, xxxx. This work was supported in part by the Postdoctoral Fellowship Program of CPSF under Grant Number GZC20261075.
        
        Yiyang Wu, Xiaohu Zhang, Yanjin Du, Tongsu Zhang, Chujun Li, Siyang Chen, Guoyi Zhang, Xiangpeng Xu are with the School of Aeronautics and Astronautics, Sun Yat-sen University, Guangzhou 510275, China. 
        (e-mail: \{wuyy239, duyj36, zhangts3, lichj29, chensy253, zhanggy57\}@mail2.sysu.edu.cn, \{zhangxiaohu, xuxp39\}@mail.sysu.edu.cn)
        \textit{(Corresponding author: Xiangpeng Xu.)}}}

\markboth{Journal of \LaTeX\ Class Files,~Vol.~14, No.~8, August~2021}%
{Shell \MakeLowercase{\textit{et al.}}: A Sample Article Using IEEEtran.cls for IEEE Journals}

\maketitle

\begin{abstract}
Accurate pose estimation is fundamental for unmanned aerial vehicle (UAV) applications, where Visual-Inertial SLAM (VI-SLAM) provides a cost-effective solution for localization and mapping. However, existing VI-SLAM methods mainly rely on sensors with limited fields of view (FoV), which can lead to drift and even failure in complex UAV scenarios. 
Although panoramic cameras provide omnidirectional perception to improve robustness, panoramic VI-SLAM and corresponding real-world datasets for UAVs remain underexplored.
To address this limitation, we first construct a real-world panoramic visual-inertial dataset covering diverse flight conditions, including varying illumination, altitudes, trajectory lengths, and motion dynamics. 
To achieve accurate and robust pose estimation under such challenging UAV scenarios, we propose a panoramic VI-SLAM framework that exploits the omnidirectional FoV via the proposed panoramic feature extraction and panoramic loop closure, enhancing feature constraints and ensuring global consistency.
Extensive experiments on both the proposed dataset and public benchmarks demonstrate that our method achieves superior accuracy, robustness, and consistency, completing all 17 sequences without tracking failure and reducing the ATE by up to 47.1\% compared with the best-performing baseline.
Moreover, deployment on an embedded platform validates its practical applicability, achieving comparable computational efficiency to PC implementations. The source code and dataset are publicly available at \href{https://github.com/YYWumm/PanoAir}{\textit{PanoAir}}
\end{abstract}


\section{Introduction}
\IEEEPARstart{V}{isual}-Inertial Simultaneous Localization and Mapping (VI-SLAM) uses cameras and an inertial measurement unit (IMU) as sensors, characterized by their compact design and cost-effectiveness. In recent years, VI-SLAM has witnessed significant advancements~\cite{campos2021orb,qin2018vins,yang2025mcnn, huang2025via}, enabling crucial applications in autonomous driving, robotics, and virtual reality. Additionally, various methods have been proposed to handle challenging lighting conditions\cite{xu2025airslam, lai2026camera}, to operate in low-texture environments \cite{si2024point, lin2024flm}, and to address camera occlusions in dynamic scenes \cite{song2022dynavins, zhang2025r}, which are common challenges faced by visual sensors.
However, SLAM using visual sensors with limited field-of-view (FOV) still struggles in such challenging environments.

One effective way to improve accuracy and robustness is to use cameras with a wide FOV \cite{zhang2016benefit}, up to a full panoramic view (360$^{\circ}$ horizontally, 180$^{\circ}$ vertically). A camera providing such a view is referred to as a panoramic camera and has attracted increasing attention in recent years\cite{lin2025one}. Various hardware solutions have been developed for panoramic imaging \cite{gao2022review}. Among them, multi-camera stitching configurations \cite{lo2021efficient} employ a small number of ultra-wide FOV fisheye lenses, often configured back-to-back, enabling an omnidirectional FOV through equirectangular projection (ERP), while maintaining a compact and lightweight design (Fig.\ref{fig:dataset}(a)). However, existing methods based on such stitching configurations either operate without metric scale\cite{sumikura2019openvslam,ji2020panoramic,huang2022360vo}, or without full SLAM to correct long-term accumulated errors \cite{wu2023360, xie2024omnidirectional}, which limits their use in practical applications. Moreover, in visually degraded conditions, such as rapid motion, low-light, or low-texture environments (Fig.~\ref{fig:dataset}(c)), existing panoramic methods often suffer from insufficient feature constraints, leading to performance degradation (Section~\ref{chap:exp}).

In addition, although numerous visual SLAM datasets have been proposed and extensively studied \cite{burri2016euroc, schubert2018tum, li2023whu}, datasets specifically for ERP images remain limited. OpenVSLAM~\cite{sumikura2019openvslam} and 360VO~\cite{huang2022360vo} datasets provide only RGB ERP images in real hand-held and simulated scenarios. The PAIR360~\cite{kim2024pair360} and 360-VIO~\cite{wu2023360} datasets provide RGB ERP images along with IMU measurements in real-world ground-vehicle and hand-held settings. AirSim360~\cite{ge2025airsim360} provides a simulated flight environment for collecting multi-modal data based on ERP images, including IMU measurements. However, despite these efforts, VI-SLAM datasets based on ERP images in real UAV flight scenarios are still absent, where the high degree of freedom and high maneuverability of UAV motion, together with sensor noise, pose significant challenges for robust visual-inertial state estimation.

To address this gap, we propose a panoramic visual-inertial SLAM framework together with a novel real-world panoramic visual-inertial UAV dataset. Specifically, we first construct a dataset collected under diverse real-world flight experiments, covering variations in flight speed, trajectory length, maneuvering patterns, and illumination (Fig.~\ref{fig:dataset}). Building upon this, we propose a panoramic visual-inertial SLAM that exploits the omnidirectional FOV of ERP images with the panoramic feature extraction and incorporates panoramic loop closure to correct accumulated drift. Finally, extensive experiments conducted on both the proposed dataset and public benchmarks, across PC and embedded platforms, demonstrate that our method achieves superior accuracy, robustness, and global consistency while maintaining comparable computational efficiency.
The main contributions are summarized as follows:

\begin{enumerate}
     \item We construct and open-source a visual-inertial SLAM dataset collected from extensive real-world UAV flight experiments. The dataset enables realistic and comprehensive evaluation of panoramic VI-SLAM systems in diverse UAV scenarios, serving as a valuable complement to existing public datasets.
    \item We propose a visual-inertial SLAM using panoramic ERP images with the proposed panoramic feature extraction and panoramic loop closure, which achieves robust, accurate, and globally consistent pose estimation.

    \item Extensive experiments on both PC and embedded platforms, using both the proposed and public datasets, demonstrate the superior performance of our method over existing approaches and validate its practical applicability.
    
\end{enumerate}

\begin{table*}[t]
\centering
\caption{Comparison of existing benchmarks for panoramic visual and visual-inertial odometry/slam}
\label{tab:tab_dataset}
\begin{tabular}{lcccccc}
\toprule
Dataset & Publication  & Setting                    & Panoramic & IMU & Real-world & Ground-truth \\ \midrule
OpenVSLAM\cite{sumikura2019openvslam} & MM'19 & Handheld                   &     \textcolor{green!70!black}{\ding{51}}      &  \textcolor{red!75!black}{\ding{55}}   &     \textcolor{green!70!black}{\ding{51}}       & \textcolor{red!75!black}{\ding{55}}             \\
Kitti-360\cite{liao2022kitti} & TPAMI'22 & Ground Vehicle             &     \textcolor{red!75!black}{\ding{55}}       &  \textcolor{green!70!black}{\ding{51}}   &       \textcolor{green!70!black}{\ding{51}}     & SFM \& GPS                       \\
360VO\cite{huang2022360vo} & ICRA'22   & Virtual scene &     \textcolor{green!70!black}{\ding{51}}      &   \textcolor{red!75!black}{\ding{55}}  &      \textcolor{red!75!black}{\ding{55}}      & Synthetic                        \\
360VIO\cite{wu2023360} & TII'23  & Ground Vehicle \& Handheld &     \textcolor{green!70!black}{\ding{51}}     &   \textcolor{green!70!black}{\ding{51}}  &     \textcolor{green!70!black}{\ding{51}}       & MoCap \& RTK                     \\
Pair360\cite{kim2024pair360} & RAL'24   & Ground Vehicle &     \textcolor{green!70!black}{\ding{51}}     &   \textcolor{orange!85!black}{\Checkmark\kern-1.2ex\raisebox{1ex}{\rotatebox[origin=c]{125}{\textbf{--}}}}  &     \textcolor{green!70!black}{\ding{51}}       & GPS                     \\
TartanAir-V2 \cite{wang2020tartanair} & IROS'25 & Virtual scene  & \textcolor{green!70!black}{\ding{51}} & \textcolor{green!70!black}{\ding{51}} & \textcolor{red!75!black}{\ding{55}} & Synthetic\\

AirSim360\cite{ge2025airsim360} & Arxiv'25 & Aerial Vehicle             &     \textcolor{green!70!black}{\ding{51}}      &   \textcolor{green!70!black}{\ding{51}}  &      \textcolor{red!75!black}{\ding{55}}      & Synthetic                        \\
RflyPano\cite{dai2026rflypano} & AAAI'26 & Aerial Vehicle & \textcolor{green!70!black}{\ding{51}} & \textcolor{green!70!black}{\ding{51}} & \textcolor{orange!85!black}{\Checkmark\kern-1.2ex\raisebox{1ex}{\rotatebox[origin=c]{125}{\textbf{--}}}} & Synthetic \\
360DVO\cite{guo2026360dvo} & RAL'26    & Mixed &     \textcolor{green!70!black}{\ding{51}}      &  \textcolor{red!75!black}{\ding{55}}   &      \textcolor{orange!85!black}{\Checkmark\kern-1.2ex\raisebox{1ex}{\rotatebox[origin=c]{125}{\textbf{--}}}}     & SFM                              \\
Ours      & - & Aerial Vehicle \& Handheld &     \textcolor{green!70!black}{\ding{51}}      &   \textcolor{green!70!black}{\ding{51}}  &     \textcolor{green!70!black}{\ding{51}}       & RTK                              \\ \bottomrule
\end{tabular}
\end{table*}

\section{Related Works}
\subsection{Wide FOV Visual SLAM}
 Visual SLAM can generally be categorized into direct \cite{engel2014lsd,engel2017direct} and feature-based \cite{qin2018vins,campos2021orb} approaches. However, when using cameras with a limited field of view (FOV), visual SLAM is prone to failure in challenging scenarios such as textureless environments or highly dynamic scenes. Expanding the FOV is a common strategy to improve robustness and accuracy as a wider view provides richer visual features. 
 
 To realize a wider FOV, existing works can be categorized into three types:
 
 (a) Single wide-FoV cameras, such as fisheye cameras \cite{matsuki2018omnidirectional, qin2018vins, campos2021orb} and annular lenses \cite{wang2022lf, ahmadi2023hdpv}, can provide a FoV exceeding 180$^{\circ}$. However, they still fail to offer full omnidirectional coverage. 
 

(b) Multi-camera systems achieve omnidirectional perception by combining multiple viewpoints, either by jointly using cameras with overlapping \cite{gao2017dual, won2020omnislam, zhang2023bamf, liu2024omninxt} or non-overlapping FoVs \cite{tribou2015multi, kaveti2023design}, or by stitching/mapping multiple images into a panoramic representation \cite{ji2020panoramic, yang2026sphere}. Nevertheless, such systems often involve multiple camera streams and require multi-camera extrinsic calibration. In contrast, a compact single-ERP panoramic camera provides a single image stream, reducing the sensing and calibration burden for lightweight UAV platforms.

\begin{figure*}[t]
\centering
\includegraphics[width=\linewidth]{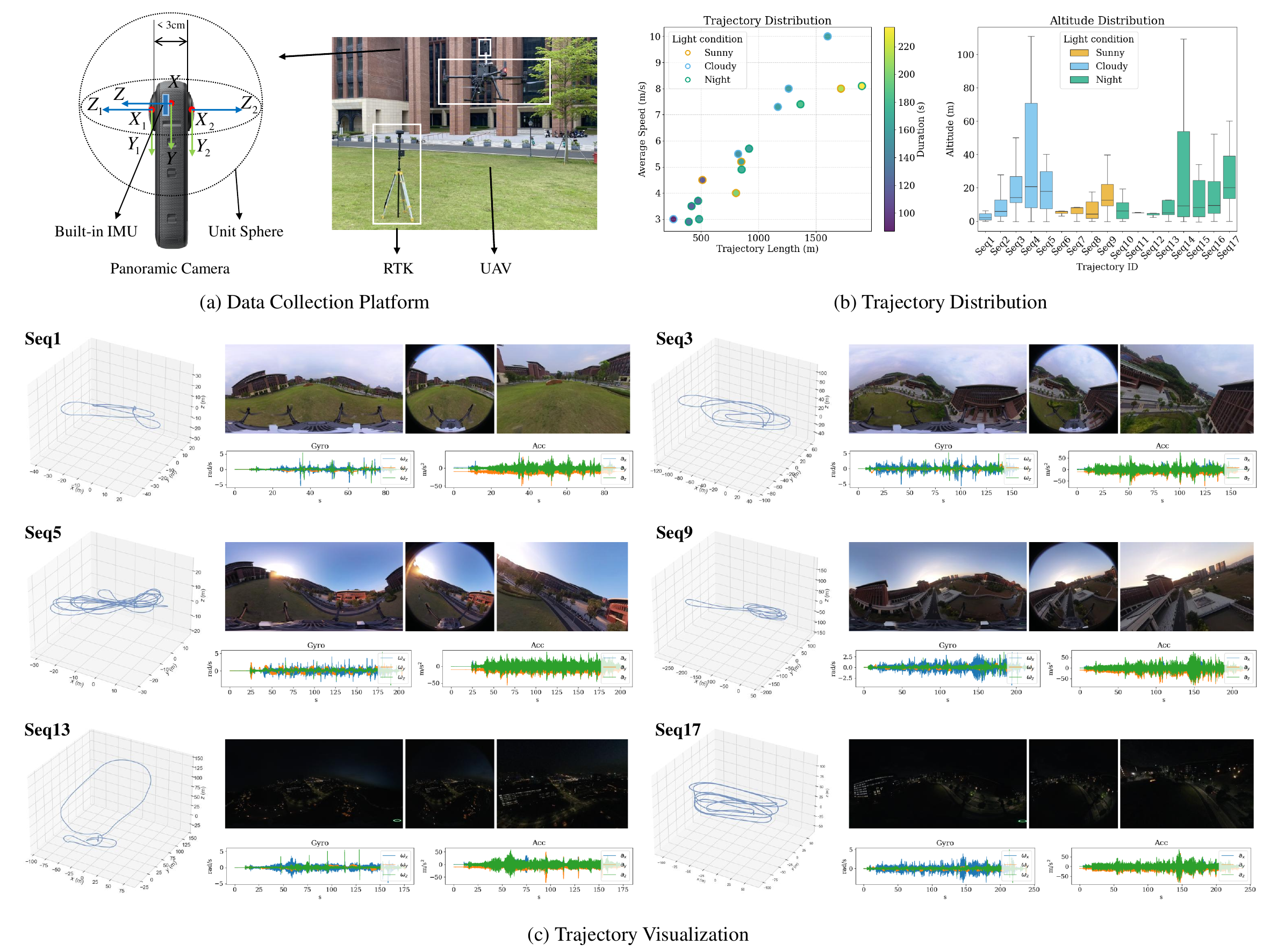}
\vspace{-20pt}

\caption{Overview of the proposed dataset. A panoramic camera is mounted on a UAV to collect panoramic-IMU data, with ground-truth trajectories recorded using an RTK ground station (a). The dataset covers diverse cross-time scenarios with varying lighting conditions, altitudes, flight speeds, and maneuvering characteristics, as summarized in (b), with representative scenes shown in (c). Further details are provided in Section\ref{sec:overview_dataset}.}
\label{fig:dataset}
\end{figure*}

(c) Compact panoramic cameras (Fig. \ref{fig:dataset}(a)) directly capture ERP images, enabling full omnidirectional coverage ($360^{\circ} \times 180^{\circ}$). OpenVSLAM \cite{sumikura2019openvslam} and 360VO \cite{huang2022360vo}, built upon \cite{mur2017orb} and \cite{matsuki2018omnidirectional}, respectively, estimate camera poses using only ERP images. However, these methods rely on monocular setups and thus cannot recover metric scale. 360-VIO \cite{wu2023360} incorporates ERP images and IMU measurements within a filter-based framework \cite{seiskari2022hybvio}. However, it lacks a complete SLAM pipeline, leading to limited accuracy and global consistency, and consequently reduced robustness under challenging conditions (Section \ref{chap:exp}).
 
 In our work, we use a compact panoramic camera to capture ERP images with a full omnidirectional FOV, together with IMU measurements, to achieve VI-SLAM within an optimization-based framework \cite{campos2021orb} by combining the panoramic optimization and panoramic loop closure.

\subsection{Deep Neural Visual SLAM}
Recent advances in deep neural networks have driven their integration into SLAM pipelines, either as replacements for individual modules or in an end-to-end manner.

Some works enhance the SLAM pipeline by selectively replacing specific modules. \cite{zhang2025real,zhao2025light} both replace traditional feature extraction and matching with SuperPoint\cite{detone2018superpoint} and Lightglue\cite{lindenberger2023lightglue}, achieving better performance 
in scenes with severe illumination changes. AirSLAM \cite{xu2025airslam} employs a unified convolutional neural network to simultaneously extract keypoints and structural lines to improve robustness. Other works directly regresses poses using deep neural end-to-end networks. DROID-SLAM\cite{teed2021droid} achieves end-to-end SLAM by incorporating BA as a differential layer, with pose and flow supervision. However, these methods are limited to pinhole images. PanoPose\cite{tu2024panopose} uses an end-to-end network to directly regress pose from panoramic image pairs, but is not suitable for real-time applications. 
360DVO\cite{guo2026360dvo} extends DPVO\cite{teed2023deep} to ERP images but operate without metric scale. 

In our work, we extract both hand-crafted and learning-based features from panoramic images, and enhance feature association with the assistance of a neural network, achieving better robustness.

\section{The Panoramic UAV Dataset}
\label{sec:overview_dataset}
Existing panoramic benchmarks \cite{sumikura2019openvslam,huang2022360vo,wu2023360,kim2024pair360,wang2020tartanair,ge2025airsim360,dai2026rflypano,guo2026360dvo} are either synthetic or collected in the real world with limited scenario coverage. In particular, there is a lack of panoramic visual-inertial datasets captured in real UAV flight scenarios, which are characterized by complex noise, distinct motion patterns, and environmental variations that are difficult to simulate. Specifically, AirSim360 \cite{ge2025airsim360} provides panoramic visual-inertial UAV data, but it is collected entirely in a simulator. 360DVO \cite{guo2026360dvo} offers real-world UAV panoramic data, yet it is captured under a stabilized viewing configuration and lacks IMU measurements.  To address this gap, we conduct real UAV experiments and construct a panoramic visual-inertial SLAM benchmark under cross-time scenarios to complement existing datasets. An overview of the proposed dataset is shown in Fig.\ref{fig:dataset}, and a comparison with existing datasets is shown in Tab.\ref{tab:tab_dataset}.

\subsection{Data Statistics}
To ensure a diverse and challenging set of evaluation scenarios, our dataset captures data under three lighting conditions: cloudy, sunny, and night, ranging from well-lit to extremely dark environments. For each lighting condition, the dataset includes a variety of trajectory lengths, flight speeds, and maneuver patterns, such as rapid yaw rotations, altitude variations, and inherent flight jitter, which are common in UAV flights. The statistics are shown in Fig.\ref{fig:dataset}(b). Our dataset consists of a total of 17 sequences, covering a cumulative distance of 15.8 km with a total duration of 45 minutes. Statistics of different sequences are shown in Tab.~\ref{tab:seqdata2}. In addition to panoramic images, our dataset also includes fisheye and pinhole images as illustrated in Fig.\ref{fig:dataset}(c). Moreover, additional handheld sequences are captured with aligned start and end positions, enabling comprehensive benchmarking of visual and visual-inertial SLAM methods (Section \ref{sec:experiment}).

\subsection{Data collection and calibration}
Our dataset is collected using an Insta360 X3~\cite{insta360x3} panoramic camera mounted on a DJI Matrice 300 RTK~\cite{djim300} UAV, capturing panoramic, front- and back-view fisheye, and pinhole images at 30 fps (Fig.~\ref{fig:dataset}(a)). The built-in IMU provides inertial measurements at 1000 Hz. The intrinsics of the dual-fisheye cameras are calibrated using Kalibr~\cite{furgale2013unified}. Pinhole images are generated by undistorting the front-view fisheye images. The extrinsic parameters between each fisheye camera and the IMU are also estimated using Kalibr. Since the stitched panoramic image does not correspond to a physical camera, its extrinsic transformation with respect to the IMU is approximated by defining a virtual camera center as the midpoint between the optical centers of the front and back fisheye cameras\cite{wu2023360}. Ground-truth trajectories are recorded at 5 Hz using a DJI D-RTK2~\cite{RTK_station} station. Sensor parameters are summarized in Tab.~\ref{tab:sequence_stats}.

\subsection{Time Synchronization}
\begin{figure}[]
  \centering
  \includegraphics[width=0.5\textwidth]{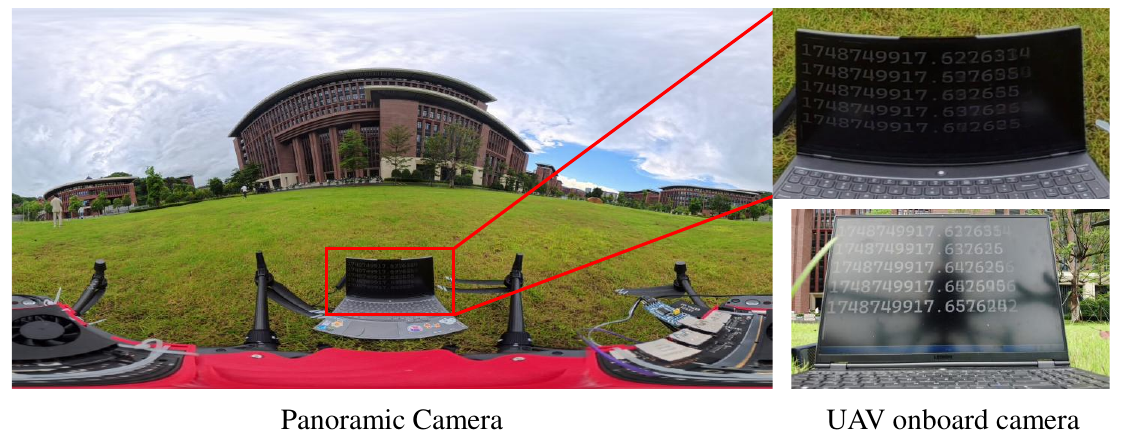}
  \vspace{-20pt}
  \caption{Time synchronization between the panoramic camera and the UAV platform.}
  \label{fig:time_sync}
\end{figure}

Since the UAV and the panoramic camera operate on independent time systems, we synchronize their timestamps using an external temporal signal similar to~\cite{kim2024pair360}. As illustrated in Fig.\ref{fig:time_sync}, the UAV onboard camera and the panoramic camera simultaneously capture the same external timer at the same frame rate. By aligning the corresponding timer readings observed in the two video streams, the temporal offset between the panoramic camera and the UAV platform can be estimated and compensated. In addition, the RTK ground station shares a consistent time reference with the UAV platform. Therefore, by parsing the SRT metadata file recorded by the UAV~\cite{djim300}, the timestamps of the panoramic camera and RTK ground truth can be synchronized within a unified temporal reference. Since the panoramic images and the measurements from the built-in IMU are internally synchronized by the panoramic camera, the corresponding IMU measurements, extracted using Gyroflow~\cite{gyroflow}, are also synchronized once the panoramic image timestamps are aligned with the UAV time system.


\begin{table}[t]
\centering
\caption{Statistics of different sequences of the proposed dataset }
\label{tab:seqdata2}
\begin{tabular}{c|c|ccc}
\toprule
\textbf{\begin{tabular}[c]{@{}c@{}}Sequence\\ \end{tabular}} & \textbf{Condition} & \textbf{\begin{tabular}[c]{@{}c@{}}Avg. Speed\\ (m/s)\end{tabular}} & \textbf{\begin{tabular}[c]{@{}c@{}}Length\\ (m)\end{tabular}} & \textbf{\begin{tabular}[c]{@{}c@{}}Duration\\ (s)\end{tabular}} \\ \midrule
Seq~1              & \multirow{5}{*}{\textbf{Cloudy}}           & 3.0                                                                 & 257                                                         & 87                                                             \\
Seq~2              &           & 5.5                                                                 & 822                                                         & 150                                                              \\
Seq~3              &           & 7.3                                                                 & 1168                                                        & 159                                                               \\
Seq~4              &           & 8.0                                                                 & 1262                                                        & 158                                                              \\
Seq~5              &           & 10.0                                                                 & 1602                                                        & 160                                                              \\ \cmidrule(lr){1-5}
Seq~6              & \multirow{4}{*}{\textbf{Sunny}}           & 4.0                                                                 & 803                                                         & 200                                                               \\
Seq~7              &          & 4.5                                                                 & 510                                                         & 114                                                              \\
Seq~8              &           & 5.2                                                                 & 849                                                        & 164                                                              \\
Seq~9              &           & 8.0                                                                 & 1718                                                         & 215                                                              \\
 \midrule
Seq~10              & \multirow{8}{*}{\textbf{Night}}           & 2.9                                                                 & 391                                                         & 133                                                           \\
Seq~11              &          & 3.0                                                                 & 481                                                         & 161                                                              \\
Seq~12              &           & 3.5                                                                 & 415                                                         & 117                                                              \\
Seq~13              &           & 3.7                                                                 & 472                                                         & 127                                                              \\
Seq~14              &          & 4.9                                                                 & 852                                                         & 172                                                              \\
Seq~15              &           & 5.7                                                                 & 917                                                         & 161                                                              \\
Seq~16              &           & 7.4                                                                 & 1365                                                        & 185                                                              \\
Seq~17              &           & 8.1                                                                 & 1901                                                        & 234                                                              \\ \bottomrule
\end{tabular}
\end{table}

\begin{table}[t]
\centering
\caption{Statistics of Sensors of the proposed dataset}
\label{tab:sequence_stats}
\resizebox{0.5\textwidth}{!}{
\large
\begin{tabular}{c|c|c}
\toprule
\textbf{Sensor} & \textbf{Parameter}                                                                                                                                                & \textbf{Rate} \\ \midrule
Camera          & \begin{tabular}[c]{@{}c@{}}Rolling Shutter\\ Panoramic: 1280x640; Front fisheye: 640x640; Pinhole: 980x640 \end{tabular}                                                             & 30 Hz         \\ \midrule
IMU             & \begin{tabular}[c]{@{}c@{}}6-axis\\ Accelerometer noise density: \SI{5.88e-4}{\meter\per\square\second\per\sqrt{\hertz}} \\
Accelerometer random walk: \SI{1.84e-4}{\meter\per\cubic\second\per\sqrt{\hertz}} \\
Gyroscope noise density: \SI{8.29e-5}{\radian\per\second\per\sqrt{\hertz}} \\
Gyroscope random walk: \SI{7.85e-5}{\radian\per\square\second\per\sqrt{\hertz}} \end{tabular} & 1000 Hz       \\ \midrule
RTK             & Horizontal: 1 cm+ 1 ppm(RMS); Vertical: 2 cm+ 1 ppm(RMS)                                                                                                            & 5 Hz          \\ \bottomrule
\end{tabular}
}
\end{table}

\begin{figure}[htbp]
  \centering
  \includegraphics[width=0.5\textwidth]{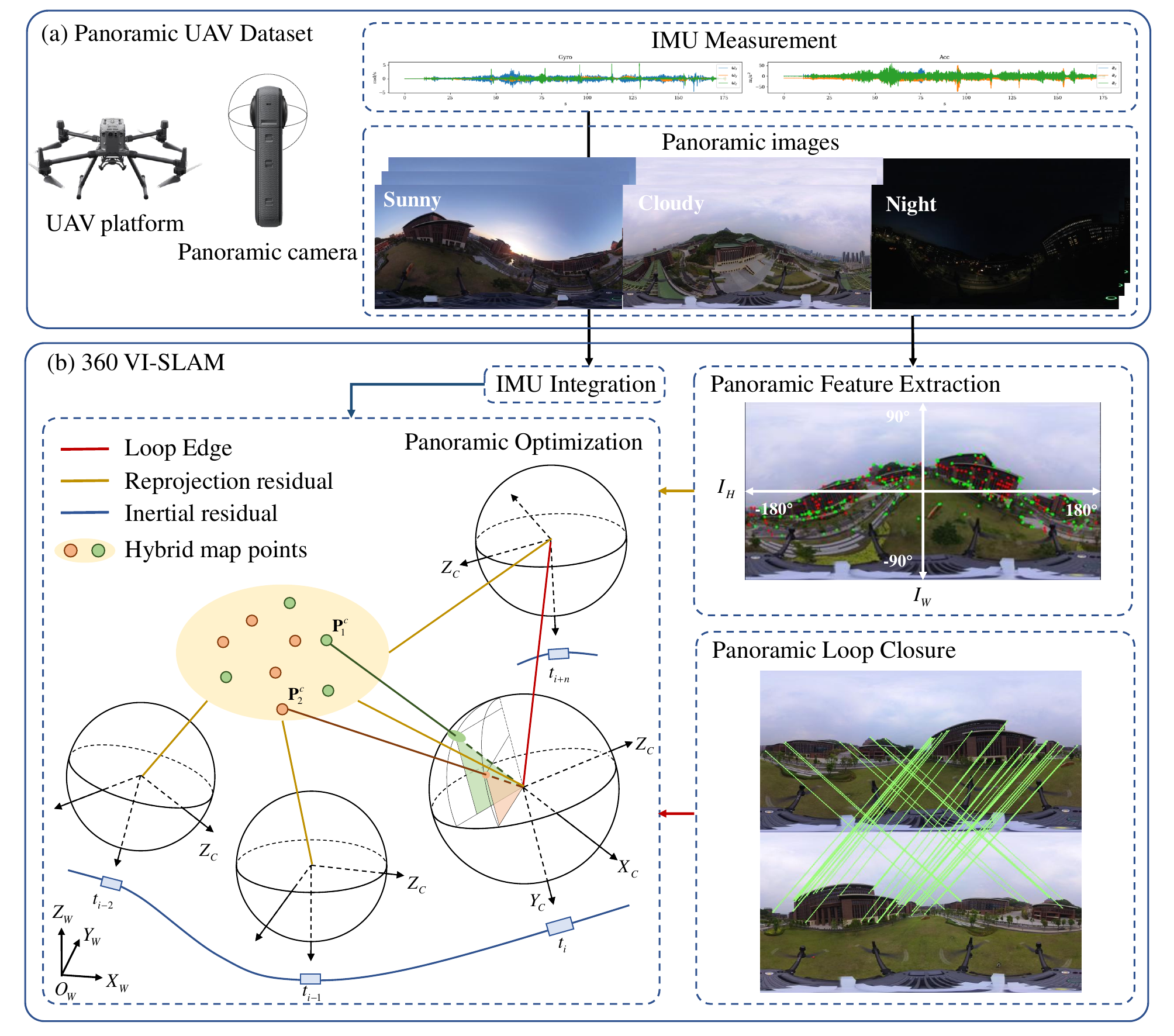}
  \vspace{-20pt}
  \caption{Overview of the proposed panoramic visual-inertial
  SLAM framework. The system takes panoramic images and IMU measurements as inputs, and achieves robust, accurate, and globally consistent results through panoramic feature extraction (Section~\ref{sec:pano_feature_extraction}), as well as panoramic optimization and panoramic loop closure (Section~\ref{sec:optimization}). Experimental results are presented in Section~\ref{chap:exp}.}
  \label{fig:proposed method}
\end{figure}
\begin{figure*}[ht]
  \centering

\includegraphics[width=\textwidth]{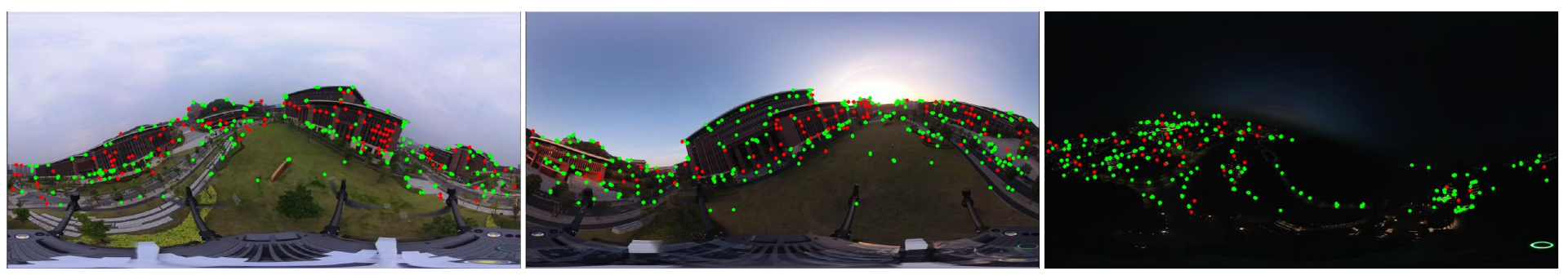}
  \vspace{-20pt}
  \caption{Hybrid feature tracking on ERP images. Red dots denote learning-based features (e.g., SuperPoint), green dots denote hand-crafted features (e.g., ORB).}

  \label{fig:show_track}
\end{figure*}

\section{Proposed}
\label{chap:sys_overview}
This section presents the proposed panoramic visual-inertial SLAM framework, consisting of panoramic feature extraction to fully exploit the omnidirectional FoV (Section~\ref{sec:pano_feature_extraction}), as well as panoramic optimization and panoramic loop closure (Section~\ref{sec:optimization}).

\subsection{Panoramic Feature Extraction}\label{sec:pano_feature_extraction}
\textbf{\textit{1) Camera model:}}
We first introduce the equirectangular projection (ERP) model, which compresses multiple perspective views into a single continuous image to establish the correspondence between 3D points and pixels on the image plane. The projection is illustrated in Fig. \ref{fig:proposed method}.


Let \( \mathbf{P}^c=[x_c, y_c, z_c]^T \) denotes the 3D point coordinates in camera frame $O_c\text{-}X_cY_cZ_c$. The ERP projection is defined as $\Pi: \mathbb{R}^3 \rightarrow \mathbb{R}^2$:
\begin{equation}
\label{equa:projection}
\mathbf{u}\!=\!\Pi(\mathbf{P}^c)
\!=\!\mathbf{K} \!
\begin{bmatrix}
\theta \\
\varphi \\
1
\end{bmatrix}\!=\!\mathbf{K} \!
\begin{bmatrix}
\operatorname{atan2}(x_c, z_c) \\
\arcsin(y_c/\sqrt{x_c^2 \!+\! y_c^2 \!+\! z_c^2}) \\
1
\end{bmatrix}
\end{equation}
where $\theta$ and $\varphi$ represent the longitude and latitude on the unit sphere, respectively, with $-\pi < \theta < \pi$ and $-\frac{\pi}{2} < \varphi < \frac{\pi}{2}$. $\boldsymbol {p}_c = [u,v]^T$ denote the pixel coordinates of 3D point on the panoramic image plane and $\mathbf{K}$ is given by
\begin{equation}
\mathbf{K}=
\begin{bmatrix}
I_W/2\pi & 0 & I_W/2 \\
0 & I_H/\pi & I_H/2
\end{bmatrix}
\end{equation}
where $I_W$ and $I_H$ denote the width and height (in pixels) of the panoramic image, respectively.

\textbf{\textit{2) Panoramic feature extraction:}}
We fully leverage the complementary strengths of hand-crafted and learning-based features to extract hybrid features from panoramic images. Although some works have proposed handcrafted features \cite{zhao2015sphorb} and learning-based features \cite{zhang2023panopoint} tailored to ERP images, they are constrained either by efficiency or limited reproducibility. 

In our work, we combine hand-crafted ORB features and learning-based SuperPoint features to balance accuracy and efficiency. Specifically, the ORB feature set is denoted as
$\mathcal{F}_{orb}=\{(\mathbf{k}^{orb}_j,\mathbf{d}^{orb}_j,\theta^{orb}_j)\}_{j=1}^{N_{orb}}$,
where $\mathbf{k}^{orb}_j$, $\mathbf{d}^{orb}_j$, and $\theta^{orb}_j$ represent the keypoint coordinate, descriptor, and orientation, respectively. The SuperPoint feature set is denoted as
$\mathcal{F}_{sp}=\{(\mathbf{k}^{sp}_j,\mathbf{d}^{sp}_j,s^{sp}_j)\}_{j=1}^{N_{sp}}$,
where $s^{sp}_j$ is the confidence score of the SuperPoint keypoint. To remove unstable learning-based features, SuperPoint features with confidence scores $s^{sp}$ lower than $\mu_{score}$ are discarded.

To avoid incorrect associations between different feature types, we adopt the strategy in~\cite{su2024hpf}. The feature type is encoded into the descriptor so that cross-type descriptor pairs are assigned the maximum distance. The Hamming distance $C(\cdot)$ of ORB and SuperPoint descriptors can be expressed as
\begin{equation}
\label{equa:class_aware_distance}
C(\mathbf{d}^{orb},\mathbf{d}^{sp}) = D,
\quad
C(\mathbf{d}^{m}_1,\mathbf{d}^{m}_2) < D,
\quad
m \in \{\mathrm{orb},\mathrm{sp}\},
\end{equation}
where $D$ is the descriptor dimension. Then, the hybrid feature sets $\mathcal{F}_{orb}$ and $\mathcal{F}_{sp}$ are fed into \cite{wang2023featurebooster} for descriptor enhancement.

Considering that ERP projection suffers from severe stretching near the top and bottom boundaries\cite{lin2025one}, which may introduce mismatches, we introduce a distortion-aware weighting function derived from the ERP projection model in Eq.~\eqref{equa:projection}. For the $j$-th feature in image $I_i$, the weight is defined as
\begin{equation}
\label{equa:uncertainty}
\sigma_{i,j}= 
\frac{
\cos\left((v_{i,j}/I_H-1/2)\pi\right)
}{\eta^l},
\end{equation}
where $v_{i,j}$ is the vertical coordinate of the feature, $I_H$ denotes the image height, $\eta$ is the pyramid scale factor, and $l$ is the pyramid level at which the feature is detected~\cite{rublee2011orb}. As features approach the top and bottom boundaries of the ERP image, $\sigma_{i,j}$ decreases, assigning lower weights to highly distorted observations and thereby reducing their influence during matching. The hybrid feature tracking result is shown in Fig.~\ref{fig:show_track}. In the representative examples shown in the left column, approximately 200 valid ORB features and 100 SuperPoint features are retained.



    

    

    

\subsection{Panoramic VI-SLAM}
\label{sec:optimization}

\textbf{\textit{1) Panoramic visual-inertial optimization:}} 
The state to be estimated at time $i$ consists of the panoramic camera pose $\mathbf{T}_i=[\mathbf{R_i},\mathbf{p_i}] \in \mathrm{SE}(3)$, its velocity $\mathbf{v}_i$ in the world frame, the gyroscope bias $\mathbf{b}_i^g$ and the accelerometer bias $\mathbf{b}_i^a$, summarized as $\mathcal{S}_i \overset{\cdot}{=} \left\{ \mathbf{T}_i, \mathbf{v}_i, \mathbf{b}^g_i, \mathbf{b}^a_i \right\}.$ Here, $\mathbf{R}_i$ and $\mathbf{p}_i$ denote the rotation and translation in world frame, and $\mathrm{SE}(3)$ denotes the special Euclidean group.

Given $k+1$ keyframes with states $\mathcal{\bar{S}}_k \overset{\cdot}{=} \{ \mathcal{S}_0, \dots, \mathcal{S}_k \}$ connected by covisibility graph \cite{campos2021orb} and $l$ 3D points with states $\mathcal{X} \overset{\cdot}{=} \{ \mathbf{x}_0, \dots, \mathbf{x}_{l-1} \}$, the panoramic visual-inertial optimization problem can be formulated as follows:

\begin{equation}
\label{equa:optimize}
    \min_{\bar{\mathcal{S}}_k,{\mathcal{X}}} \left( 
\sum_{i=1}^{k} \left\| \mathbf{r}_{\mathcal{I}_{i-1,i}} \right\|_{\Sigma^{-1}_{\mathcal{I}_{i,i+1}}}^2 
+ \sum_{j=0}^{l-1} \sum_{i \in \mathcal{K}^j} \rho_{\text{Hub}}\left( \left\| \mathbf{r}_{ij} \right\|_{\Sigma^{-1}_{ij}} \right) 
\right)
\end{equation}
where $\mathbf{r}_{\mathcal{I}{i-1,i}}$ denotes the inertial residual with IMU integration formulated in \cite{campos2021orb}. 
$\mathbf{r}_{ij}$ denotes the reprojection error of the observed map point $j$ on the image plane of frame $i$, with pixel coordinate defined in Eq. \eqref{equa:projection}. 
\begin{equation}
\label{equa:residual}
    \mathbf{r}_{ij}=\mathbf{u}_{ij}-\Pi(\mathbf{T}_i^{-1}\mathbf{P}_j^c)
\end{equation}
The set $\mathcal{K}^j$ consists of keyframes that observe the 3D point $j$. The robust Huber kernel $\rho_{\text{Hub}}$ is employed to mitigate the influence of outliers. 
The distortion introduced by panoramic images is explicitly taken into account in the definition of the covariance matrix \(\Sigma_{ij}\) in Eq. \eqref{equa:optimize}, where it is defined as follows:
\begin{equation}
    \Sigma_{ij}^{-1} =  \sigma^2_{i,j} \begin{bmatrix} 1 & 0 \\ 0 & 1 \end{bmatrix}
\end{equation}
$\sigma_{i,j}$ is given by Eq. \eqref{equa:uncertainty}.

Eq. \eqref{equa:optimize} is solved using the Gauss-Newton method with the Jacobian matrix derived as follows:
\begin{equation}
\begin{aligned}
\label{equa:Jacobi}
    \mathbf{F}_{ij} &= \frac{\partial \mathbf{r}_{ij}}{\partial\mathbf{P}^{c}_{ij}}\frac{\partial\mathbf{P}^{c}_{ij}}{\partial \delta \boldsymbol{\xi_i}}, &&
    \mathbf{E}_{ij} = \frac{\partial \mathbf{r}_{ij}}{\partial\mathbf{P}^{c}_{ij}}\frac{\partial\mathbf{P}^{c}_{ij}}{\partial\mathbf{P}^{w}_{ij}}
\end{aligned}
\end{equation}

where the camera pose $\boldsymbol{\xi}_i \in \mathfrak{se}(3)$ and the map point in world frame $\mathbf{P}^{w}_{ij}$ are update iteratively:
\begin{equation}
\begin{aligned}
\label{equa:Jacobi}
    \mathbf{J}_{ij}^T\Sigma_{ij}^{-1}\mathbf{J}_{ij} \begin{bmatrix} \delta \boldsymbol{\xi_i}   \\ \delta \mathbf{{P}}^{w}_{ij}  \end{bmatrix}  &= -\mathbf{J}_{ij}^T\Sigma_{ij}^{-1}\mathbf{r}_{ij} , && \mathbf{J}_{ij}= [ \mathbf{F}_{ij} \; \mathbf{E}_{ij}]
\end{aligned}
\end{equation}

\textbf{\textit{2) Panoramic Loop Closure:}} 
Loop closure corrects accumulated drift during long-term SLAM, ensuring global consistency, which is not considered in previous methods \cite{huang2022360vo,wu2023360, guo2026360dvo}. We adopt a coarse-to-fine strategy for panoramic loop closure:

\textit{(1) Coarse loop keyframe candidate detection:}
For each new panoramic keyframe $K_i$ inserted into the map, coarse loop candidate detection is first performed. A hybrid bag-of-words (BoW) vector is constructed from the ORB and SuperPoint descriptors and used to query the DBoW2 \cite{GalvezTRO12} database. The three keyframes with the highest similarity scores are retrieved as loop candidates from the candidate set $\mathcal{S}$, which contains all map keyframes except the $N_\mathrm{neighbor}=50$ keyframes adjacent to $K_i$.

\textit{(2) Geometric verification:}
Once the coarse loop candidates are obtained, each candidate undergoes a series of geometric verification steps. First, hybrid BoW-based descriptor matching is performed between $K_i$ and the candidate keyframes. If a sufficient number of matches (at least 50 matches) is obtained, the relative $\mathrm{SE(3)}$ transformation is then estimated using RANSAC and subsequently refined with robust outlier rejection. Finally, the estimated transformation is further verified through projection matching (at least 80 matches) and a gravity-direction consistency check (less than $1^\circ$). Specifically, under the ERP projection model in Eq.~\eqref{equa:projection}, map points may lie in the negative-$z$ half-space, as shown in Fig.~\ref{fig:proposed method}, which can still be validly observed and projected onto the panoramic image. Therefore, to fully exploit the omnidirectional FoV of panoramic images, we omit the conventional positive-depth check commonly used in limited-FoV camera settings during geometric verification. This enables accurate $\mathrm{SE(3)}$ estimation even when the two keyframes have opposite viewing directions, as illustrated by Loop1 in Fig.~\ref{fig_quality_results}.

\textit{(3) Loop correction:}
The estimated $\mathrm{SE(3)}$ transformation between $K_i$ and the loop keyframe is then propagated through pose graph optimization and further refined via global bundle adjustment to complete loop correction, both considering ERP distortion as in Eq.~\eqref{equa:optimize}.
Loop closure detection and the experimental results are shown in Fig.~\ref{fig_quality_results}, Tab.\ref{tab:loopcount} and Section~\ref{chap:exp}.

\begin{table*}[]
\caption{Tracking accuracy on our proposed panoramic UAV benchmark (ATE,in meters)} 
\label{tab:compare_on_our_dataset}
\centering
\resizebox{\textwidth}{!}{
\begin{tabular}{lccccccccccccccccccccc}
\toprule
\multirow{2}{*}{Method} & 

\multicolumn{2}{c}{Settings}&
\multicolumn{5}{c}{Cloudy}                                                                                             & \multicolumn{4}{c}{Sunny}                                                                  & \multicolumn{8}{c}{Night} 
 & \multirow{2}{*}{Avg.}
 & \multirow{2}{*}{Succ.}                                                                                                                                                    \\ \cmidrule(lr){2-3} \cmidrule(lr){4-8} \cmidrule(lr){9-12} \cmidrule(lr){13-20} 
                                & Sensor
                                & LC & Seq1                       & Seq2                 & Seq3                 & Seq4                 & Seq5                 & Seq6                 & Seq7                 & Seq8                 & Seq9                 & Seq10                & Seq11                & Seq12                & Seq13                & Seq14                & Seq15                & Seq16                & Seq17                \\ \midrule
ORB-SLAM3\cite{campos2021orb}                       & 
FI & \ding{51} &

0.800                      & 0.596                & \ding{55}                    & 3.969                & \ding{55}                    & 0.403                & 0.293                & 0.633                & 1.874                & 3.503                & 0.907                & 1.019                & 0.868                & 4.539                & 3.030                & \ding{55}                    & 1.637   & \ding{55}     & 83\%         \\
VINS-mono\cite{qin2018vins}                      & 

FI & \ding{51} &

0.555                      & 1.085                & 1.889                & 2.835                & 2.816                & \cellcolor{yellow!40}0.359               & 0.344                & 1.153                & 3.417                & \ding{55}                    & \ding{55}                    & \ding{55}                    & \ding{55}                    & \ding{55}                    & \ding{55}                    & \ding{55}                    & 10.930  &\ding{55} & 59\%              \\ 
Droid-SLAM\textsuperscript{*}\cite{teed2021droid}   & 
P& \textemdash &

\cellcolor{red!30}\textbf{0.243} & \cellcolor{yellow!40}0.369       & \cellcolor{yellow!40}0.435 & \ding{55} & \cellcolor{yellow!40}1.211 & 0.423 & 0.253 & 11.387 & \cellcolor{red!30}\textbf{1.046} & \cellcolor{yellow!40}0.519 & 3.970 & \cellcolor{red!30}\textbf{0.675} & \cellcolor{yellow!40}0.373 & \ding{55} & \ding{55} & \ding{55} & 15.425  & \ding{55} & 76\% \\ \midrule
OpenVSLAM\textsuperscript{*}\cite{sumikura2019openvslam}                       
& O & \ding{51}
&0.370                      & 0.535                & 0.496                & \cellcolor{red!30}\textbf{0.703}                & \cellcolor{red!30}\textbf{0.758}                & 0.722                & 0.449                & \cellcolor{yellow!40}0.544                & 1.294                & \cellcolor{red!30}\textbf{0.478}                & \cellcolor{yellow!40}0.864                & \ding{55}                    & 0.593                & \cellcolor{red!30}\textbf{0.382}                & \cellcolor{red!30}\textbf{0.755}                & \cellcolor{red!30}\textbf{0.799}                & \cellcolor{yellow!40}1.397   & \ding{55}    & 94\%          \\
360DVO\textsuperscript{*}\cite{guo2026360dvo} &
O & \ding{55} &
0.388 & 1.199 & 1.119 & 1.677 & 3.677 & 1.259 & \cellcolor{yellow!40}0.247 & 0.893 & 3.927 & 1.044 & 0.978 & \cellcolor{yellow!40}0.696 & \cellcolor{red!30}\textbf{0.298} & 0.828 & 0.984 & 2.582  &1.963 & \cellcolor{yellow!40}1.398 & 100\%
\\
360-VIO\cite{wu2023360}                         
& OI & \ding{55}
&4.912                      & 4.282                & 3.513                & 13.587               & 11.867               & 1.501                & 1.804                & 4.343                & 9.585                & \ding{55}                    & \ding{55}                    & 10.675               & \ding{55}                    & \ding{55}                    & \ding{55}                    & \ding{55}                    & \ding{55}  & \ding{55}   & 59\%                \\
OpenVINS\cite{geneva2020openvins} & 
OI & \ding{55} &
1.384 & 1.618 & 3.012 & 9.772 & 13.900 & 1.300 & 0.403 & 2.452 & 5.171 & 2.082 & \ding{55} & 10.419 & \ding{55} & 19.701 & 4.148 & 6.613 & 6.926  & \ding{55} & 88\%
\\
\textbf{Ours}                            
&OI & \ding{51}
& \cellcolor{yellow!40}0.300                      & \cellcolor{red!30}\textbf{0.265}                & \cellcolor{red!30}\textbf{0.433}                & \cellcolor{yellow!40}0.824                & 2.204                & \cellcolor{red!30}\textbf{0.343}                & \cellcolor{red!30}\textbf{0.239}                & \cellcolor{red!30}\textbf{0.438}                & \cellcolor{yellow!40}1.205                & 0.933                & \cellcolor{red!30}\textbf{0.690}                & 0.838                & 0.476                & \cellcolor{yellow!40}0.500                & \cellcolor{yellow!40}0.875                & \cellcolor{yellow!40}0.827                & \cellcolor{red!30}\textbf{1.170}         
&
\cellcolor{red!30}\textbf{0.739} & 
100\%      
\\
\bottomrule
\end{tabular}
}
\begin{tablenotes}
\footnotesize
\item{Best results and second-best results are highlighted as \colorbox{red!30}{first} and \colorbox{yellow!40}{second}. '\ding{55}' indicates failure. '*' indicates scale-aligned with ground truth. }
\item{F= Fisheye, P= Pinhole, O= Omnidirectional, I= IMU, LC= LoopClosure. DROID-SLAM uses dense global BA instead of LC.}
\end{tablenotes}
\end{table*}

\begin{table}[t]
\centering
\caption{Repeated-run statistics for failed cases in Table IV}
\label{tab:failed_run_statistics}

\resizebox{\columnwidth}{!}{
\begin{tabular}{c c c c}
\hline
\textbf{Method} & 
\textbf{Sequence} & 
\textbf{Successful Runs} & 
\textbf{ATE RMSE (m)} \\
\hline

\multirow{3}{*}{ORB-SLAM3~\cite{campos2021orb}}
& Seq3  & 2/5 & $4.129 \pm 0.122$ \\
& Seq5  & 1/5 & $4.383$ \\
& Seq16 & 2/5 & $1.463 \pm 0.283$ \\
\hline

\multirow{7}{*}{VINS-Mono~\cite{qin2018vins}}
& Seq10 & 0/5 & -- \\
& Seq11 & 0/5 & -- \\
& Seq12 & 0/5 & -- \\
& Seq13 & 0/5 & -- \\
& Seq14 & 1/5 & $10.677$ \\
& Seq15 & 0/5 & -- \\
& Seq16 & 0/5 & -- \\
\hline

\multirow{4}{*}{DROID-SLAM~\cite{teed2021droid}}
& Seq4  & 0/5 & -- \\
& Seq14 & 0/5 & -- \\
& Seq15 & 0/5 & -- \\
& Seq16 & 0/5 & -- \\
\hline

OpenVSLAM~\cite{sumikura2019openvslam}
& Seq12 & 0/5 & -- \\
\hline

\multirow{7}{*}{360-VIO~\cite{wu2023360}}
& Seq10 & 0/5 & -- \\
& Seq11 & 0/5 & -- \\
& Seq13 & 0/5 & -- \\
& Seq14 & 0/5 & -- \\
& Seq15 & 0/5 & -- \\
& Seq16 & 0/5 & -- \\
& Seq17 & 0/5 & -- \\
\hline

\multirow{2}{*}{OpenVINS~\cite{geneva2020openvins}}
& Seq11 & 0/5 & -- \\
& Seq13 & 0/5 & -- \\
\hline

\end{tabular}
}

\vspace{1mm}
\begin{minipage}{\columnwidth}
\footnotesize
\textit ATE RMSE is reported as mean $\pm$ standard deviation
over successful runs. For a single successful run, only the RMSE is
reported; ``--'' indicates no successful run.
\end{minipage}
\end{table}

\section{Experiments}
Extensive experiments are conducted to demonstrate that: (1) the omnidirectional FoV and the proposed panoramic feature extraction provide stronger feature constraints and improve performance; (2) the proposed panoramic loop closure corrects
accumulated drift and improves global consistency; and (3) the evaluation of our panoramic VI-SLAM on an embedded platform demonstrates its feasibility for onboard UAV deployment.

\label{chap:exp}

\subsection{Experiment Setup}
\textbf{\textit{1) Dataset:}}
\label{sec:implement detail}
Currently, there are very few publicly accessible and usable real-world panoramic visual-inertial SLAM datasets. We conducted comparative experiments on two  real-world visual-inertial SLAM datasets, both of which include panoramic images with IMU measurements: (a) Our panoramic dataset, captured during UAV flight scenarios, consists of a total of 17 sequences under three conditions (\textit{Cloudy, Sunny, and Night}), along with two additional handheld sequences with aligned start and end points (Section \ref{sec:experiment}); (b) The three available sequences of the 360-VIO dataset \cite{wu2023360} collected in handheld scenarios with \textit{Easy, Medium, and Hard}. 

\begin{figure*}[t]
  \centering
  \includegraphics[width=\textwidth]{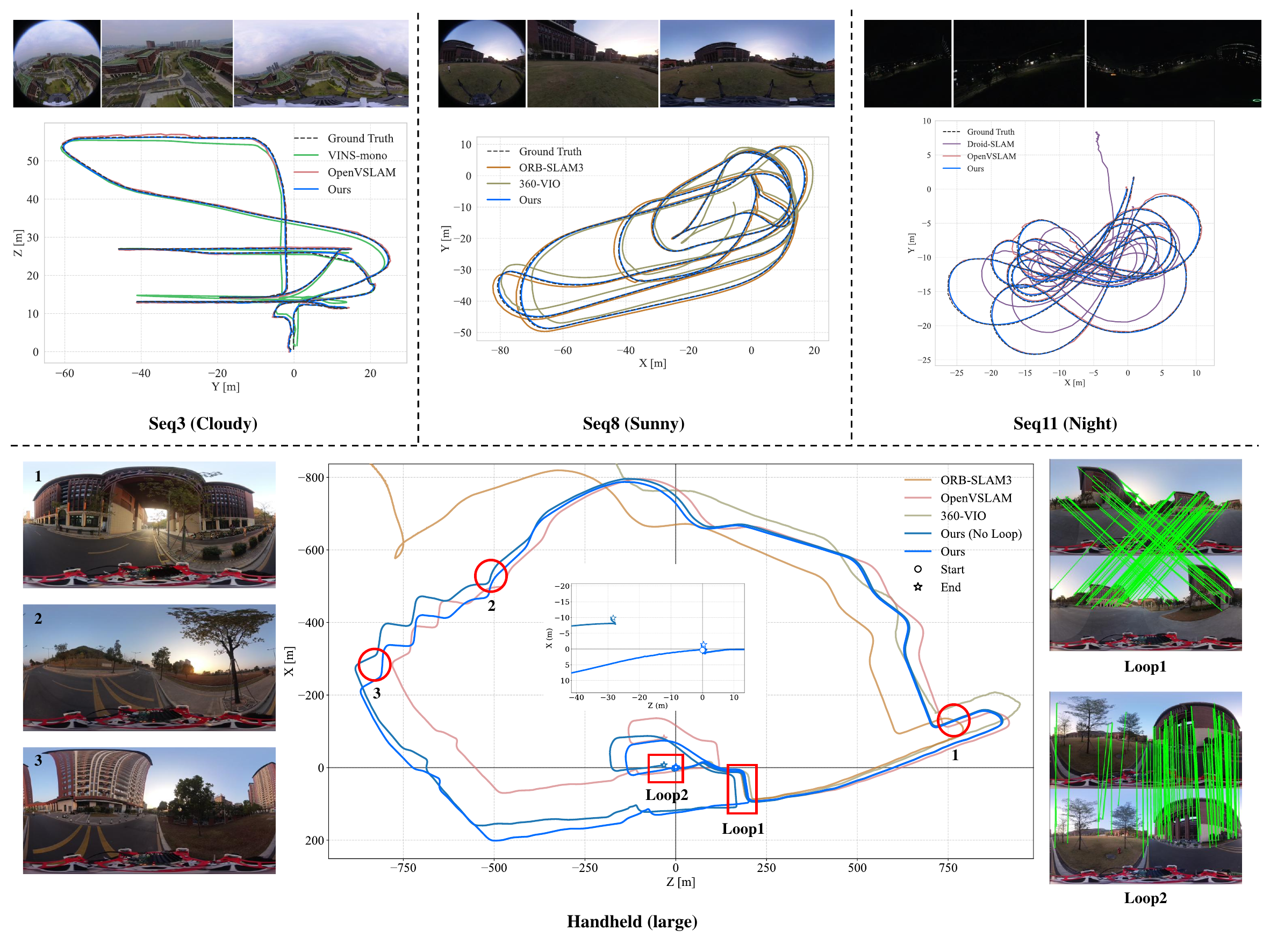}
  \vspace{-22pt}
  \caption{Qualitative comparison results. 
Top: comparison of different methods on the proposed dataset, where our method produces estimates closer to the ground truth and demonstrates stronger robustness. 
Bottom: results on an additional large-scale handheld sequence. 
Other methods suffer from varying degrees of drift, while our method achieves lower drift and, by leveraging panoramic loop closure, successfully aligns with the starting position, resulting in a globally consistent estimation.
}
  \label{fig_quality_results}
\end{figure*}

\textbf{\textit{2) Compared Methods:}}
Our method was evaluated against state-of-the-art approaches from three categories: (a) \textit{Pinhole Learning-based Visual SLAM}: DROID-SLAM \cite{teed2021droid}; (b) \textit{Pinhole/Fisheye Classic Visual-Inertial SLAM}: ORB-SLAM3 \cite{campos2021orb} and VINS-Mono \cite{qin2018vins}; (c) \textit{Panoramic Visual/Visual-inertial SLAM/Odometry}: OpenVSLAM \cite{sumikura2019openvslam}, OpenVINS\cite{geneva2020openvins}, 360-VIO \cite{wu2023360} and 360 DVO\cite{guo2026360dvo}.
Each method was run five times per sequence, and the mean translational RMSE of ATE/RPE (m) was reported using evo \cite{grupp2017evo}. Scale alignment was applied only to monocular-only trajectories, while other methods were evaluated without scale correction.
Following the commonly adopted evaluation criteria in \cite{zhang2025real,guo2026360dvo,wu2023360,qiu2025mac, xu2025airslam}, a method was marked as failed if any run failed to initialize, remained in a tracking-lost state for more than half of the frames in a sequence, entered the subsequent relocalization state, or showed significant trajectory drift (e.g., an ATE greater than 30 m), as exemplified by the ORB-SLAM3 trajectory in the bottom row of Fig. \ref{fig_quality_results}.



\textbf{\textit{3) Implementation details:}}
All experiments were conducted on an Intel i9-14900K CPU, 64GB RAM, and NVIDIA RTX 4090 D GPU, running Ubuntu 22.04. Our framework was implemented in C++, with deep networks deployed via ONNX Runtime. 
In our hybrid feature extraction, the ORB extractor was set to 2000 keypoints, while SuperPoint features with $\mu_{score}<0.1$ were discarded. The input settings and loop-closing configurations of all methods are shown in Tab.~\ref{tab:compare_on_our_dataset} and Tab.~\ref{tab:acc_360vio_ours}, respectively. Furthermore, to assess the performance of our method in embedded deployment, we deployed it on an embedded NVIDIA Jetson Orin NX platform with 16GB RAM\cite{nvidia_jetson_orin}, using TensorRT for edge-device evaluation.

\begin{table*}[t]
\centering
\caption{Tracking accuracy on 360-VIO benchmark (ATE and RPE, in meters)}
\label{tab:acc_360vio_ours}
\resizebox{0.65\textwidth}{!}{
\begin{tabular}{lcccccccccc}
\toprule
\multirow{2}{*}{Methods} 
&\multicolumn{2}{c}{Settings}
&\multicolumn{2}{c}{Seq 1 (Easy)}      & \multicolumn{2}{c}{Seq 2 (Medium)}            & \multicolumn{2}{c}{Seq 3 (Hard)}  &   \multicolumn{2}{c}{Avg.}      \\ \cmidrule(l){2-3}
\cmidrule(l){4-5} \cmidrule(l){6-7} \cmidrule(l){8-9} \cmidrule(l){10-11}
& Sensor & LC & ATE & RPE & ATE & RPE & ATE & RPE & ATE & RPE
                                \\ \midrule
ORB-SLAM3\cite{campos2021orb}$^\dagger$
& FI & \ding{51}
& 0.186                & 0.108                & 0.256                & 0.068                & \colorbox{red!30}{0.186}                & \colorbox{red!30}{0.086}   & \colorbox{yellow!40}{0.209} &     {0.087}          \\
VINS-mono\cite{qin2018vins}$^\dagger$ 
& FI & \ding{51}
& 1.406                & 0.082                & 2.960                & 0.242                & \ding{55}                & \ding{55} &\ding{55} & \ding{55}                \\
OpenVSLAM\textsuperscript{*} \cite{sumikura2019openvslam} 
& O & \ding{55}           
& \colorbox{red!30}{0.149}                & \colorbox{yellow!40}{0.049}              & \colorbox{red!30}{0.058}                &  \colorbox{yellow!40}{0.068}            & \ding{55}                & \ding{55}   & \ding{55} & \ding{55}             \\ 

360DVO\textsuperscript{*} \cite{guo2026360dvo} 
& O & \ding{55}
& 0.184 &  \colorbox{red!30}{0.049} & 0.217 & \colorbox{red!30}{0.067} & 0.520 & \colorbox{yellow!40}{0.089} & 0.307 & \colorbox{red!30}{0.068}

\\
360-VIO\cite{wu2023360}$^\dagger$
& OI & \ding{55}   
& 0.456                & 0.058                & 0.390                & 0.088                & 0.632                & 0.122   & 0.493 &  0.089           \\
\textbf{Ours (w/o loop)}     
& OI & \ding{55}
& \colorbox{yellow!40}{0.154} & 0.050 & \colorbox{yellow!40}{0.064} & {0.070} & \colorbox{yellow!40}{0.218} & {0.099} & \colorbox{red!30}{0.145} & \colorbox{yellow!40}{0.073} \\ \bottomrule
\end{tabular}
}
\begin{tablenotes}
\footnotesize
\centering
\item{Best results and second-best results are highlighted as \colorbox{red!30}{first} and \colorbox{yellow!40}{second}.'\ding{55}' indicates failure. '*' indicates scale-aligned with ground truth.}
\item{F= Fisheye, P= Pinhole, O= Omnidirectional, I= IMU, LC= LoopClosure.}

\item{$^\dagger$ Results are taken from the original 360-VIO\cite{wu2023360}.}

\end{tablenotes}
\end{table*}

\subsection{Comparison}
\label{sec:experiment}
\textit{\textbf{1) Tracking accuracy:}}
The tracking results on the proposed panoramic visual-inertial UAV dataset are shown in Tab.~\ref{tab:compare_on_our_dataset}, with additional repeated-run statistics for the failed cases provided in Tab.~\ref{tab:failed_run_statistics}. Our method completes all 17 sequences without tracking failure and demonstrates strong overall accuracy, reducing the ATE by up to 47.1\% compared with the best-performing baseline 360DVO, while providing poses with metric scale. In the more challenging \textit{Night} scenarios  with low-light and high-agility motion (e.g. \textit{Seq12, Seq14, Seq16}), some methods suffer from initialization failure or tracking loss. The performance of our method can be attributed to the omnidirectional FOV of panoramic camera providing richer visual information with more observed landmark compared to the fisheye or pinhole method like ORB-SLAM3 and Droid-SLAM. 

Specifically, the high accuracy of DROID-SLAM on some easy sequences (e.g., \textit{Seq1, Seq7}) benefits from post-processing global bundle adjustment, whereas the limited pinhole FOV leads to significant front-end drift on more challenging sequences (e.g., \textit{Cloudy Seq5, Sunny Seq8, Night Seq17}). 360DVO achieves relatively high accuracy and robustness even with only monocular odometry. However, it still underperforms compared to OpenVSLAM on some sequences. Although OpenVSLAM achieves high accuracy on some sequences, its lack of true metric scale and reliance solely on visual information limit its applicability. In contrast, our method integrates IMU measurements and leverages panoramic hybrid feature extraction to provide richer feature constraints, enabling robust tracking even under sparse or degraded visual conditions. Moreover, 360-VIO using only odometry suffers from drift and global inconsistency. And its optical-flow-based tracking degrades under extreme low-light \textit{Night} scenarios which VINS-mono also encounter. We observed that OpenVINS (with binocular mode) generally achieves better performance than 360-VIO but still suffers from drift and global inconsistency. In contrast, our method extracts hybrid features with robust matching and corrects drift through panoramic loop closure, achieving high accuracy and global consistency.

The tracking results on panoramic visual-inertial public benchmark \cite{wu2023360} are shown in Tab. \ref{tab:acc_360vio_ours}, with some results taken directly from the paper. To enable a fair comparison with visual-inertial odometry methods, we \textbf{disable the loop closure module} in our approach. The results show that our method achieves the highest success rate and accuracy across three indoor handheld sequences, reducing the average ATE by 30.6\% compared with ORB-SLAM3 and by 70.6\% compared with 360-VIO. 360DVO also achieves comparably high accuracy, especially in RPE metrics, but falls short in terms of global consistency. OpenVSLAM suffers from tracking failure on the \textit{hard} sequence due to the limited robustness of using pure RGB information, while VINS-Mono is constrained by its FOV and also fails on the \textit{hard} sequence. As discussed in \cite{wu2023360}, ORB-SLAM3 attains high accuracy when loop closure (relocalization) is enabled, but failures occur when these modules are disabled.

\textit{\textbf{2) Qualitative results:}}
The results on our proposed panoramic visual-inertial UAV Dataset are shown in the upper part of Fig.\ref{fig_quality_results}. Our method is closer to the ground truth across different scenarios, demonstrating superior accuracy and robustness. To further evaluate our method and to enrich the diversity of the panoramic visual-inertial dataset in real-world settings, we additionally collected two handheld trajectories with aligned start and end points for evaluation. The first trajectory \textit{Indoor handheld medium} is approximately 330 m long with a duration of about 230 seconds. The second trajectory \textit{Outdoor handheld large} is approximately 5.5 km in length and lasts about 880 seconds. Both trajectories contain dynamic objects and are affected by illumination variations. The settings are identical to those in Tab. \ref{tab:sequence_stats}. 

The results of the second trajectory \textit{Outdoor handheld large} are shown in the lower part of Fig.\ref{fig_quality_results}. The end-to-end learning-based method DROID-SLAM ran out of GPU memory.  ORB-SLAM3 and 360-VIO exhibit noticeable drift, while OpenVSLAM, even with loop detection enabled, fails to correct drift induced by monocular scale variations. In contrast, our method achieves the smallest drift without loop closure and successfully aligns the start and end positions when loop closure is enabled. 
The end positions are reported in Tab.~\ref{tab:loop_endposition}, and the results show that the proposed panoramic loop closure effectively corrects accumulated drift.

Notably, benefiting from the omnidirectional FOV, our panoramic loop closure detects loops not only from the same viewing direction (Loop 2), but also from opposite directions (Loop 1), which is not achievable with pinhole or fisheye cameras. 
To further quantify this capability, we compare the number of detected loops with and without the positive-depth check described in Section~\ref{sec:optimization}. As shown in Tab.~\ref{tab:loopcount}, removing the positive-depth check and exploiting the full omnidirectional FoV enables more valid loop closures to be detected, particularly for revisited locations observed from opposite viewing directions. Besides, the ablation results are presented in Tab.~\ref{tab:ablation}. As shown in Tab.~\ref{tab:ablation}, enabling the proposed panoramic loop closure consistently improves the ATE over the corresponding results without loop closure, with no accuracy degradation observed in the evaluated sequences.

\begin{table}[]
\centering
\caption{End positions of two handheld sequences with aligned start and end positions.
}
\label{tab:loop_endposition}
\begin{tabular}{lcc}
\toprule
\multicolumn{1}{c}{\multirow{2}{*}{\begin{tabular}[c]{@{}c@{}}End position\\ (x,y,z in meters)\end{tabular}}} & \multirow{2}{*}{\begin{tabular}[c]{@{}c@{}}Handheld indoor\\ 330 m\end{tabular}} & \multirow{2}{*}{\begin{tabular}[c]{@{}c@{}}Handheld outdoor\\ 5.5 km\end{tabular}} \\
\multicolumn{1}{c}{}                                                                                          &                                                                                  &                                                                                    \\ \midrule
Ours(Without Loop)                                                                                            & -1.52, -0.53, -0.74                                                                & -9.92, -49.37, -28.43                                                               \\
Ours(With Loop)                                                                                               & \textbf{-0.01, 0.00, 0.01}                                                                  & \textbf{-1.39, -0.08, 0.36}                                                                   \\ \bottomrule
\end{tabular}
\end{table}

\begin{table}[t]
\caption{Detected loop closure counts with and without positive-depth check}
\resizebox{0.5\textwidth}{!}{
\begin{tabular}{lccccc}
\toprule
Setting & Cloudy & Sunny & Night & Indoor & Total count \\
\midrule
With positive-depth check
& 18 & 20 & 16 & 10 & 64 \\

\textbf{Without positive-depth check}
& \textbf{21} & \textbf{23} & \textbf{17} & \textbf{11} & \textbf{72}  \\

\bottomrule
\end{tabular}
\label{tab:loopcount}
}
\end{table}

\begin{figure*}[t]
    \centering
    \includegraphics[width=\linewidth]{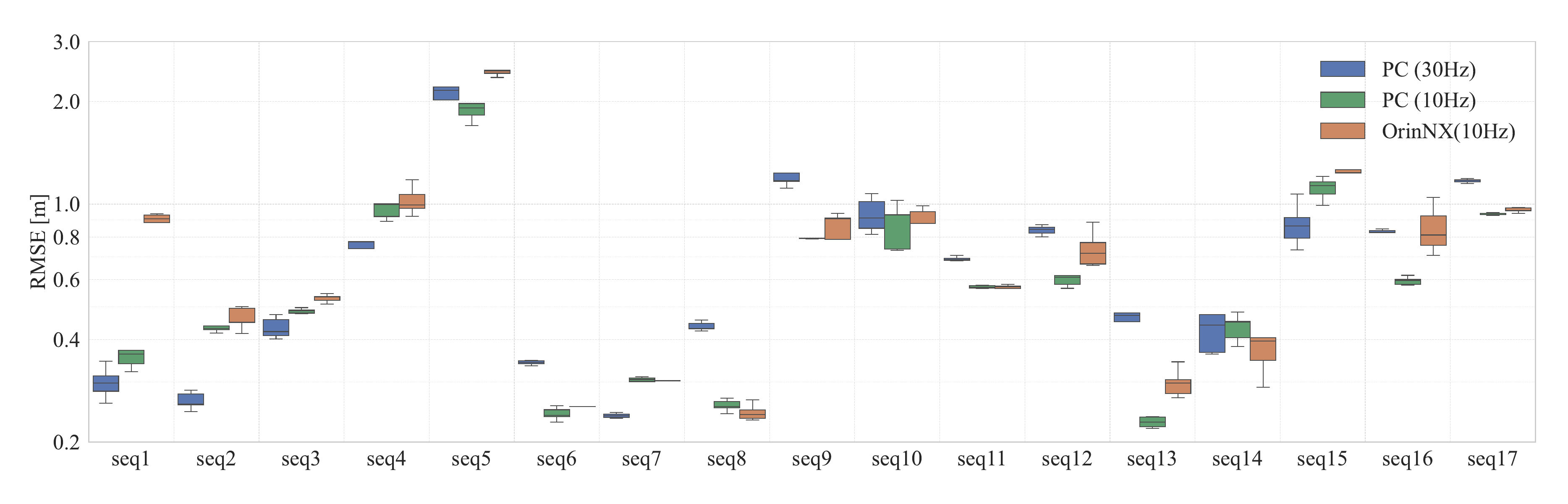}
\vspace{-20pt}
\caption{Box plots of ATE over five runs on the PC with 10 Hz and 30 Hz frame inputs, and on the Orin NX with 10 Hz frame input. The error variations remain stable across different platforms and input frequencies, demonstrating the feasibility of onboard deployment.
} \label{fig:boxplot}
\end{figure*}



\subsection{Embedded deployment}

To evaluate the performance of our proposed method on a computationally constrained platform, we deploy it on a Jetson Orin NX edge device and accelerate the hybrid feature extraction using ONNX and TensorRT. We report results under three input frame-rate settings: 30 Hz and 10 Hz on the PC platform, and 10 Hz on the Orin NX platform, while all other configurations remain unchanged.

Fig.~\ref{fig:boxplot} shows the box plots of ATE for all sequences on different platforms, demonstrating comparable accuracy between the PC and the Jetson Orin NX at different input frequencies. The runtime statistics are evaluated on three long trajectories from the proposed dataset (\textit{Cloudy (Seq5), Sunny (Seq9), Night (Seq17)}) and reported in Tab. \ref{tab:runtime}. Specifically, “Hybrid Feature Extraction” in Tab.~\ref{tab:runtime} refers to the complete Panoramic Feature Extraction module described in Section \ref{sec:pano_feature_extraction}. Our method achieves high-accuracy pose estimation on the edge device at a comparable processing speed, while maintaining accuracy comparable to PC implementation. In real-world deployments, high-rate pose outputs can be obtained by fusing IMU integration with SLAM pose corrections, making the system suitable for UAVs and other embedded platforms. Such capabilities are generally difficult to achieve with monocular-based approaches like\cite{sumikura2019openvslam,guo2026360dvo}.

\begin{table}[t]
\caption{Ablation results of the proposed module. Averages computed over all sequences in different scenarios.}
\resizebox{0.5\textwidth}{!}{
\begin{tabular}{lccccc}
\toprule
\multirow{2}{*}{Method}    & \multicolumn{5}{c}{ATE(m)$\downarrow$}                                                         \\ \cmidrule(l){2-6}
                           & Cloudy         & Sunny          & Night          & Indoor         & Avg.           \\ \midrule
w/o learning-based feature         & 1.095          & 0.558 & 0.930          &          \ding{55}      & \ding{55}               \\
w/o hand-crafted feature & 1.566
& \textbf{0.553} & \ding{55} & 0.574 & \ding{55} \\

w/o distortion weight & 1.342 & 0.556 & 0.789 & 0.150 & 0.709

\\
w/o panoramic loop closure & 2.170          & 1.087          & 1.517          & \textbf{0.145} & 1.223          \\
w/o omnidirectional FOV & 1.482 & 0.599 & \ding{55} & 0.322 & \ding{55} \\
\textbf{Ours(Full System)}              & \textbf{0.805} & 0.566          & \textbf{0.789} & 0.151          & \textbf{0.578} \\
\bottomrule
\end{tabular}
\label{tab:ablation}
}
\end{table}

\begin{table*}[h]
\caption{Runtime statistics of different platforms evaluated on three long trajectories from the proposed dataset.}
\label{tab:runtime}
\centering
\begin{tabular}{lcccccccccccc}
\toprule
\multirow{2}{*}{Runtime(ms)} & \multicolumn{3}{c}{Cloudy(Seq5)} & \multicolumn{3}{c}{Sunny(Seq9)} & \multicolumn{3}{c}{Night(Seq17)} & \multicolumn{3}{c}{Avg.} \\ \cmidrule(l){2-4} \cmidrule(l){5-7} \cmidrule(l){8-10} \cmidrule(l){11-13} 
                             & PC\textsuperscript{1}        & PC\textsuperscript{2} 
                             & OrinNX\textsuperscript{3}    & PC\textsuperscript{1}        & PC\textsuperscript{2} 
                             & OrinNX\textsuperscript{3}    
                             & PC\textsuperscript{1}        & PC\textsuperscript{2} 
                             & OrinNX\textsuperscript{3} 
                             & PC\textsuperscript{1}        & PC\textsuperscript{2} 
                             & OrinNX\textsuperscript{3}    \\ \midrule
HybridFeature Extraction     & 14.0        & 14.1 & 76.5               & 13.9       & 13.6 & 74.3             & 12.4 & 12.7      & 65.6              & 13.4 & 13.5  & 72.1           \\
Local Optimization           & 9.0 &  6.9      & 32.8               & 9.2 & 7.0      & 33.1               & 7.9 &  6.4     & 29.7              & 8.7 &  6.8   & 31.9           \\ \midrule
Total Tracking               & 25.6 & 23.1        & 118.7              & 26.1  & 23.3     & 119.8              & 22.9     & 20.8 & 104.0             & \textbf{24.9(40fps)}  & \textbf{22.4(45fps)}   & \textbf{114.2(9fps)}         \\ \bottomrule
\end{tabular}
\begin{tablenotes}
\footnotesize
\item{PC\textsuperscript{1}:30Hz input, PC\textsuperscript{2}:10Hz input, OrinNX\textsuperscript{3}:10Hz input}
\end{tablenotes}
\end{table*}

\subsection{Ablation Study}
\label{sec:ablation}
The ablation study focuses on two modules: panoramic feature extraction and panoramic loop closure (Fig.~\ref{fig:proposed method}). 
We evaluate the hybrid features by replacing them with ORB-only extraction (w/o learning-based feature), SuperPoint-only extraction (w/o hand-crafted feature) and by removing the distortion-aware weights (w/o distortion weight). The panoramic loop closure is assessed by disabling it (w/o panoramic loop closure). The omnidirectional FoV is assessed by masking half of each ERP image, retaining only a 180$^\circ$ FoV (w/o omnidirectional FoV). This masking-based ablation is intended to isolate the effect of FoV reduction and is not strictly equivalent to a real single-fisheye camera, which has a different distortion model and effective resolution distribution. Results on the proposed dataset (\textit{Cloudy, Sunny, Night}) and public dataset \cite{wu2023360}(\textit{Indoor}) are shown in Tab. \ref{tab:ablation}. 

Among the ablated variants, w/o panoramic loop closure exhibits worse overall trajectory accuracy, indicating that panoramic loop closure effectively corrects drift. The w/o learning-based feature, w/o hand-crafted feature, and w/o omnidirectional FoV variants provide insufficient visual constraints, resulting in tracking failures or degraded accuracy. The w/o distortion weight variant also shows degraded overall accuracy. These results validate the effectiveness of the proposed modules.

\section{Conclusion}
This paper presents a robust and accurate panoramic visual-inertial SLAM, together with a new real-world panoramic visual-inertial UAV dataset. 
Our dataset comprises 17 sequences collected in real-world UAV flight scenarios, covering varying illumination conditions, diverse motion patterns, altitude distributions, and velocity profiles, and also includes two handheld sequences in both medium-scale and large-scale environments, thereby providing a valuable complement to existing SLAM datasets.

Our proposed panoramic visual-inertial SLAM with improvements benefit from the omnidirectional field of view (FOV) of panoramic cameras, together with IMU measurements, as well as the panoramic feature extraction and panoramic loop closure. It completes all the proposed 17 sequences without tracking failure and reducing the ATE by up to 47.1\% compared with the best-performing baseline. Embedded deployment on Jetson Orin NX further shows that our method maintains reasonable computational efficiency, offering a high-accuracy pose estimation solution using panoramic visual-inertial sensors for UAVs and other robotic platforms.
 
Nevertheless, our method still has several limitations. 
First, the overall framework can be further accelerated to better exploit the abundant GPU resources available on modern edge devices~\cite{khabiri2025fasttrack}. Second, although the omnidirectional FoV improves robustness to occlusions and dynamic objects, these factors are not explicitly modeled, and large-scale occlusions or highly dynamic scenes may still cause erroneous pose estimation and mapping. Future work will focus on panoramic perception for UAV platforms\cite{li2024robust} and high-quality panoramic mapping.

\bibliographystyle{IEEEtran}
\bibliography{ref}

\end{document}